\documentclass[conference]{IEEEtran}
\IEEEoverridecommandlockouts
\usepackage{amsmath,amssymb,amsfonts}
\usepackage{algorithmic}
\usepackage{graphicx}
\usepackage{textcomp}
\usepackage{xcolor}
\usepackage{booktabs}
\usepackage[
backend=biber,
style=ieee,
sorting=ynt
]{biblatex}

\usepackage{fancyhdr}

\begin{document}

\title{Detrive: Imitation Learning with Transformer Detection for End-to-End Autonomous Driving
}

\author{
\IEEEauthorblockN{1\textsuperscript{st} Daoming Chen}  
\IEEEauthorblockA{\textit{Bristol Robotics Laboratory} \\
\textit{University of Bristol}\\
Bristol, UK \\  
ta21463@bristol.ac.uk}

\\

\IEEEauthorblockN{3\textsuperscript{rd} Feng Chen}
\IEEEauthorblockA{\textit{Zhejiang VIE Science \& Technology Co. Ltd.} \\
Zhejiang, China \\  
mervyn@vie.com.cn}

\and 

\IEEEauthorblockN{2\textsuperscript{nd} Ning Wang}
\IEEEauthorblockA{\textit{Bristol Robotics Laboratory } \\
\textit{University of the West of England }\\
Bristol, UK \\
ning2.wang@uwe.ac.uk}

\\

\IEEEauthorblockN{4\textsuperscript{th} Tony Pipe} 
\IEEEauthorblockA{\textit{Bristol Robotics Laboratory} \\
\textit{University of the West of England}\\
Bristol, UK \\
Anthony.Pipe@uwe.ac.uk}
}

\maketitle

\thispagestyle{fancy}            
\fancyhead{}                     
\fancyfoot[l]{979-8-3503-4353-3/23/\$31.00 ©2023 IEEE}
\fancyfoot[c]{\quad}                    
\renewcommand{\headrulewidth}{0pt}      
\renewcommand{\footrulewidth}{0pt}
\pagestyle{empty}                

\begin{abstract}
This Paper proposes a novel Transformer-based end- to-end autonomous driving model named Detrive. This model solves the problem that the past end-to-end models cannot detect the position and size of traffic participants. Detrive uses an end-to-end transformer based detection model as its perception module; a multi-layer perceptron as its feature fusion network; a recurrent neural network with gate recurrent unit for path planning; and two controllers for the vehicle's forward speed and turning angle. The model is trained with an on-line imitation learning method. In order to obtain a better training set, a rein- forcement learning agent that can directly obtain a ground truth bird's-eye view map from the Carla simulator as a perceptual output, is used as teacher for the imitation learning. The trained model is tested on the Carla's autonomous driving benchmark. The results show that the Transformer detector based end-to- end model has obvious advantages in dynamic obstacle avoidance compared with the traditional classifier based end-to-end model.
\end{abstract}


\section{Introduction}
Since 1988, the first autonomous vehicle, ALVINN, has been recognised as a successful example of the use of neural networks for autonomous driving \cite{pomerleau1988alvinn}. Modern autonomous driving technology was started since 2004, as DARPA \cite{darpa} hosted its first grand challenge in self-driving, and attracted the brightest minds for studying this topic. After 2007, with the appearance of Lidar, high-resolution sensors, and the success of the DARPA Urban Challenge, the foundation of autonomous driving technology was built. Companies like Google (Waymo) started their research in autonomous vehicles (AV), and the so-called AV 1.0 was formed.

One of the most important technology revolutions for au- tonomous driving is the rise of Deep Learning in 2012. Since then, the AV commercial companies have flourished around the world. In the early stages, Artificial Intelligence (AI) was applied only to computer vision in the AV industry. In 2021, the concept of AV 2.0 was stated by a company called Wayve \cite{hawke2021reimagining}. AV 2.0 means an end-to-end autonomous driving approach that uses only one large neural network for all driving challenges. It replaces the old control-based decision making module with a neural network. This is the key to solving the scale-up problem that AVs face now.

One of the biggest breakthroughs in AI in the last five years is the module called transformer \cite{vaswani2017attention}. Since its emergence, almost all models in the Natural Language Process (NLP) field have been inseparable from the transformer module. In recent years, researchers start to explore the possibility of applying transformer in the computer vision field \cite{dosovitskiy2020image} \cite{liu2021swin} \cite{carion2020end} \cite{zhang2022dino}, and those transformer based models are breaking the limitation of traditional CNN based models. However, there are not enough studies of the application of transformers in autonomous driving.

Most previous end-to-end autonomous driving methods only use a classification model as their perception module \cite{codevilla2018end} \cite{chen2015deepdriving}
\cite{sauer2018conditional} \cite{chen2020learning} \cite{chitta2023transfuser} \cite{wu2022trajectory}. One of the reasons why detection models like Faster-RCNN \cite{ren2015faster} or YOLO \cite{redmon2016you} are not used in end- to-end AD models is that neither of those models are end-to- end. Therefore, to solve this problem, the end-to-end detection model, DETR \cite{carion2020end}, is used in this paper's proposed model, which has not been studied before.

Motivated by the problem mentioned above, this paper aims to explore the application of using transformer models as the perception module for an end-to-end autonomous driving. The model is developed based on the open source CARLA simu- lator \cite{dosovitskiy2017carla}, and the model is evaluated on the CARLA leader- board \cite{carla} to fairly compare with other people's methods. The source code of the model used on CARLA leaderboard is released on https://github.com/Alexbeast-CN/Detrive

\section{RELATED WORK}
Since the 1980s, more and more researchers have been working  on  autonomous  vehicles  using  the  end-to-end paradigm \cite{pomerleau1988alvinn}. This long-history paradigm now has two dif- ferent branches. One branch is based on deep reinforcement learning and another branch is based on imitation learning. The deep reinforcement learning method is to allow the robot coverage to an optimal policy \cite{sutton2018reinforcement}. The imitation learning method uses an expert to make labels for the deep neural network to learn from \cite{argall2009survey}.

The advantage of the end-to-end paradigm is that using neural networks is easier to scale up comparing to those hand- crafted, rule-based conventional methods. Autonomous driving is a typical problem affected by the long tail effect. There are countless corner cases in everyday driving scenarios which are not covered by the training dataset. The modular pipeline models have to firstly define those cases and then hand code the solution. On the other hand, the AI-based end-to-end model can solve these problems by collecting data which is much easier \cite{hawke2021reimagining}.

However, the deep learning methods are still a black box to human. When an end-to-end AV made a wrong decision and caused an accident, it's difficult to explain how the decision was made and how to prevent such problems in the future \cite{tampuu2020survey}.

\subsection{Intermediate representation}
The intermediate representation paradigm was first proposed in 2015 \cite{chen2015deepdriving}. It is a hybrid version of the end-to-end paradigm and the pipeline paradigm. It emphasises using state-of-the- art deep learning for the perception module to obtain more comprehensive and accurate state observations. These state observations are also known as affordances. As defined in \cite{sauer2018conditional}, the affordances are objects or rules that temporally or spatially limited the movement of the ego vehicle. The items of affordances are listed as follows:

\begin{itemize}
  \item The distance from the ego vehicle's centre point to the centre line of the lane $d$.
  \item The relative angle between the ego vehicle's head direction and the centre line of the lane $\psi$.
  \item The distance from the ego vehicle to the front vehicle $l_1$.
  \item The distance from the ego vehicle to a pedestrian $l_2$.
  \item The traffic lights.
  \item The speed sign.
\end{itemize}

In recent years, this paradigm has been widely accepted by industry and academia, but in a bird's-eye-view (BEV) form instead \cite{tesla} \cite{hu2021fiery} \cite{li2022bevformer} \cite{liu2023bevfusion}. The BEV intermediate representation is to map the sensor input into a top down view. The first step is to use a neural network to detect the major traffic participants and estimate their relative position to the ego vehicle. Then, a sensor fusion algorithm will transform the features into a BEV map.

\subsection{Conditional Imitation Learning}

One of the breakthroughs for autonomous driving via imita- tion learning is proposed by Codevilla et al \cite{codevilla2018end} \cite{codevilla2019exploring}. Compared to previous methods that only use images as input, Conditional Imitation Learning (CIL) embeds other sensor input such as speed and the navigational command. The embedding
method uses a multilayer perceptron neural network (MLP) to transform the raw data into the same dimension as the output image features. Those image features together with other feature is then feed into another MLP to generate actions. The architecture of the model is shown in Figure 1. With these auxiliary inputs, the CIL model is able to drive the vehicle in many complex scenarios. Therefore, it is one of the strongest baselines for end-to-end autonomous driving. Until today, the architecture of conditional imitation learning can still be found in the state of the art models \cite{wu2022trajectory}.

\begin{figure}[htbp]
    \centering
    \includegraphics[width=0.5\textwidth]{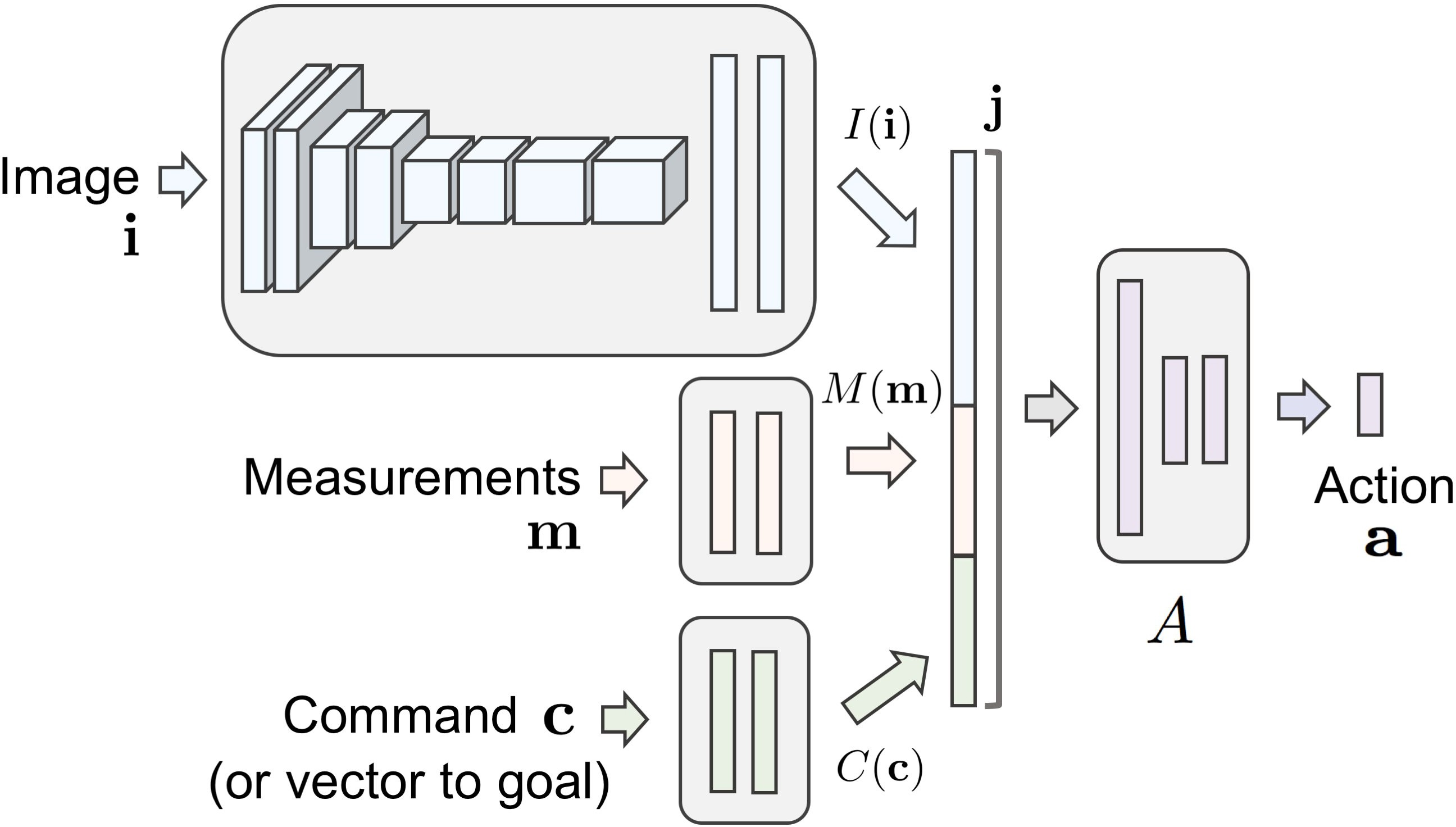}
    \caption{The architecture of conditional imitation learning\cite{codevilla2018end}}
    \label{fig:1}
\end{figure}

\section{METHODOLOGY}

\subsection{System Overview}

Previous end-to-end imitation learning approach, such as CILRS \cite{codevilla2018end} is able to drive in urban streets, but it is very likely to collide with other traffic participants in dense traffic sce- narios. One of the possible reasons for this defect comes from the perception module. The perception module in CILRS is an ImageNet pretrained ResNet \cite{he2016deep}, which can only classify images into 1000 object labels. However, this doesn't make sense for autonomous driving, since the location and shape of each object can't be detected by ResNet. Therefore, to improve driving ability, the classification perception module needs to be replaced by a detection model. The detailed illustration of this model is in the Section III-B.

Except for the architecture of the imitation learning network, the dataset for training the model is another key component for good results. Previous methods normally use a human as an expert to generate datasets. However, to update a neural network with over a million parameters will take the expert thousands of hours of demonstration. It is not only time- consuming, but also a waste of financial investment and human effort. Another way to collect an expert dataset is to hand code an autopilot based on the traffic rules. Those autopilots can access the ground truth information from the simulator. In most cases, the autopilot can drive the vehicle safely. However, it is very difficult to hand code an rule-based autopilot that can perfectly handle all driving scenarios. To tackle this problem, this paper uses a reinforcement learning agent which is trained based on the ground truth Bird's-Eye-View (BEV) map. The RL expert is much more robust than the rule-based autopilot when evaluated on the NoCrash-dense benchmark \cite{zhang2021end}.

\begin{figure}[htbp]
    \centering
    \includegraphics[width=0.5\textwidth]{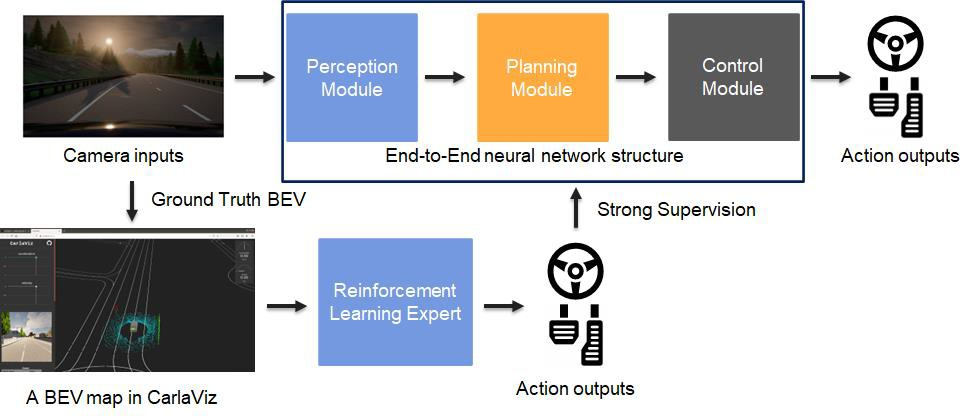}
    \caption{The overall architecture of the Detrive model}
    \label{fig:2}
\end{figure}

The model proposed in this paper is called Detrive. It is an autonomous driving model that uses DETR as its perception module. The overall architecture of the Detrive model is shown in Figure 2. There are two models, the RL model uses no sensor but a ground truth BEV map for generating actions that form the training dataset. The imitation learning model is an agent that can be deployed on an ego vehicle. There are three main modules in the imitation learning model, the perception module can detect both the label and bounding box of an object; the planning module takes the detection result as input and generates future waypoints; the control module use those waypoints to generate driving actions. The inputs of the ego vehicle are a front view monocular camera, speedometer, and a Global Navigation Satellite System (GNSS) that can give a global navigation from point A to point B.

The imitation learning model is another key component of this paper. The overview of the model is shown in Figure 3. As mentioned above, the model can be divided into three main modules. The perception module is modified from the DETR model. It uses a backbone to extract features from the input images. The extracted features will then be positional encoded and pass through a transformer module. The transformer module will transform those features into a vector of object classification and bounding boxes. The vector then will be concatenated with tensors from other sensors. The concate- nated vector is the input of the trajectory planning network. The planning module first uses a multilayer perceptron (MLP) to perform information fusion on the input vector. Then this vector is passed together with the goal point tensor, into a gated recurrent unit (GRU) network for predicting future way points. Lastly, the control module uses a PID controller to generate steering angle, throttle, and brake action to fit the predicted trajectory.

\subsection{Perception Module}

The perception module uses a DETR-like network architec- ture that is divided into four parts, a backbone for extracted features from image input; a positional encoding to encode feature maps with positional information; transformer encoder and decoder for object detection.

\subsubsection{Backbone}
DETR has no special requirements for backbone, any neural network that can do image classification can be used as its backbone. However, the efficiency of the network and the prediction accuracy needs to be considered before deploying DETR to a task. Due to the real-time requirements of the automatic driving system and the limitation of available computing power, the backbone can only choose a medium-sized classification network. After some testing, ResNet50 \cite{he2016deep} was finally selected as the backbone of this model. The backbone takes a list of images \(\{x_{1}\), \(x_{2}\),..., \(x_{n}\}\) as its input where \(x_{i}\in\mathrm{R}^{3\times H_{0}\times W_{0}}\). The output of the backbone is a list of feature maps \(\{f_{1}\), \(f_{2}\),..., \(f_{n}\}\), where \(f_{i}\in\mathrm{R}^{c\times H\times W}\), \(c=2048\), \(H=H_{0}/32\), \(W=W_{0}/32\). In this work, the size of the input image is \(256\times 256\)
\subsubsection{Positional encoding}
In order to understand the context of a full picture, the Transformer's attention mechanism shuffles the ordering of the input signal. Therefore the feature maps passed into the Transformer must be positional encoded to output meaningful results.

\subsubsection{Transformer Encoder}
After the backbone, the dimension of the feature map needs to be reduced by a \(1\times 1\) convolutional layer to form a new feature map \(z\in\mathrm{R}^{d\times H\times W}\). In this model \(d=256\). Because the transformer encoder only takes two dimensions, the third-dimension tensor \(z\) needs to be flattened to generate the input vector \(z\in\mathrm{R}^{d\times HW}\). The last process for the transformer encoder is to positional encode the vector \(z\).

\subsubsection{Transformer Decoder}
The decoder also has the same structure as the original transformer. It uses the output from encoder as the input keys and values of the multi-head self-attention layer. However, the difference is that it uses **N** objects as the queries of the multi-head self-attention layer. Then, the decoder will transform the object queries into a latent layer that contains the fused information of that object. The latent layer can be further decoded into confidence of that object's label and a bounding box of the object's position by a Feed Forward Network (FFN). The structure of the FFN is very simple. It is a network with three fully connected layers with Rectified Linear Unit (ReLU) as the activation function between layers. There are two different FFNs in DETR, one is used to predict the object labels, another one is for predicting the bounding box.

\subsection{Planning Module}

\begin{figure*}[ht]
    \centering
    \includegraphics[width=\textwidth]{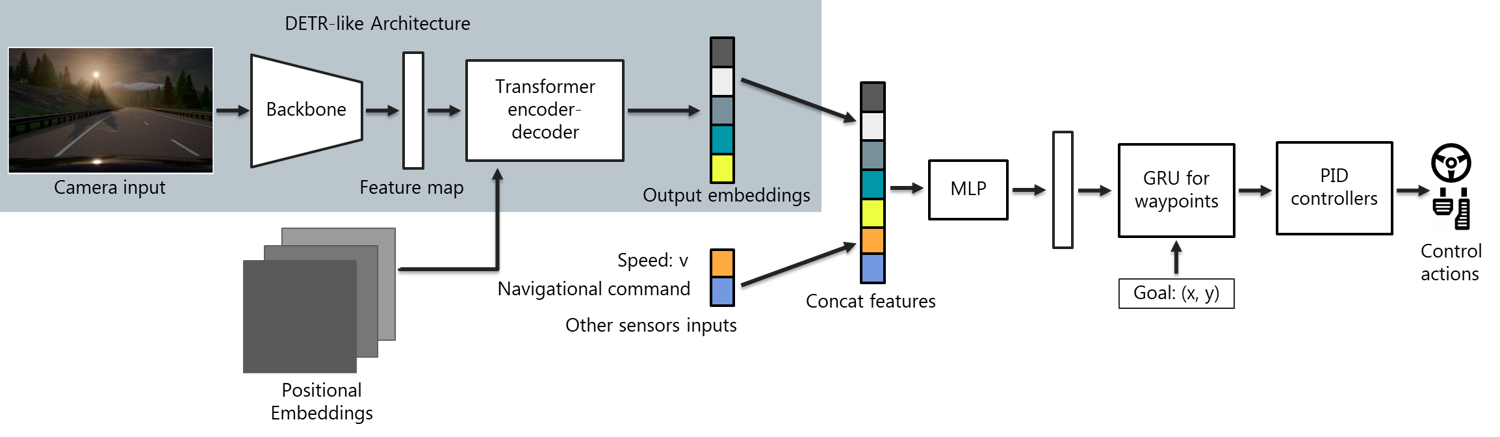}
    \caption{ The architecture of the proposed Detrive model}
    \label{fig:3}
\end{figure*}

The planning module consists of a feature fusion network and a Recurrent Neural Network (RNN) with Gated Recurrent Units (GRU) Cells. The planning module is also known as the decision module. It takes the output of the perception network as input, and generates dynamic trajectory waypoints. This predicted trajectory is updated at a high frequency to allow the vehicle to cope with dynamic obstacles and traffic signals.

\subsubsection{Feature Fusion}
The feature fusion network is used to fuse the perception outputs with other sensor data. It provides a scene understanding ability for the agent by combining perception modules with other sensor information. The structure of the fusion network is shown in Figure 4. The outputs of the perception module are object labels and object bounding box. As shown in the figure, the shape of the outputs are respectively \((b,\,N,\,1)\) and \((b,\,N,\,4)\), where \(b\) is the batch size, \(N\) is the number of objects. In order to reduce the feature loss caused by the deepening of the network, a residual structure is used. It will pass the feature map from the backbone to a MLP then concatenate with object label and bounding box tensors. The last two dimensions of the concatenation tensor will be flattened into one dimension. The the shape of the new tensor is \((b,\,(5\times N))\). Afterwards, the new tensor will be fed into a single-layer MLP that can fuse the data and transform the shape into \((b,d)\).

\begin{figure}[htbp]
    \centering
    \includegraphics[width=0.5\textwidth]{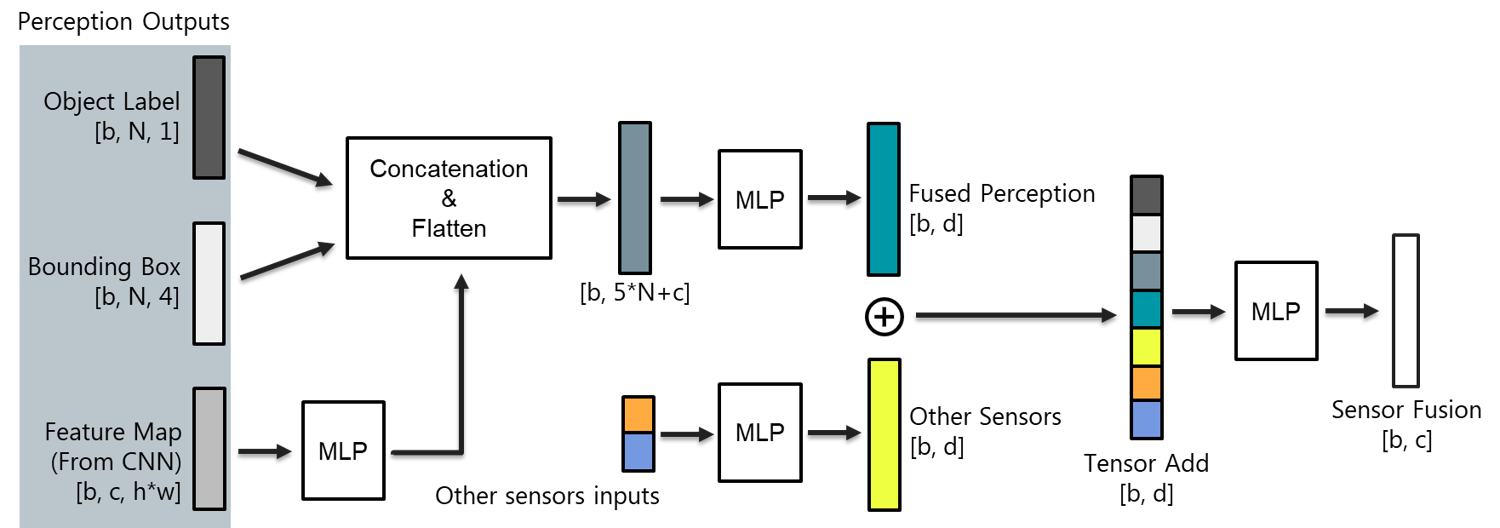}
    \caption{The architecture of the feature fusion network}
    \label{fig:4}
\end{figure}

Navigational command and speed are the other two inputs of the model in addition to the RGB camera. There are six kinds of navigation commands in GNSS, they are to move along the current lane, change to the left lane, change to the right lane, turn left at the intersection, turn right at the intersection, and go straight at the intersection. These six commands will be one-hot encoded into numbers from 0 to 5. Then, the navigational command and speed will be transformed into tensors and fed into a similar MLP to generate vector with the shape of \((b,d)\). After that, the two tensors will be added element-wise to form a new feature tensor. Finally, another three-layer MLP will further fuse the information and reduce tensor dimension to \((b,c)\).

\subsubsection{Recurrent Neural Network}

The recurrent neural network (RNN) is the most important part of the planning module. Compared with CNN, RNN is better at dealing with sequence problems. The trajectory prediction problem is similar to the one-to-many or many-to-many model in the sequence problem. In RNN, \(F\) is the parametric network, \(h\) and \(x\) are the inputs of the network. These three elements form the basic cell of an RNN. It can be formalized as Equation 1, where \(W\) is the learnable parameters matrix, \(h\) is the hidden output, and \(x\) is the input. More specifically, \(h_{0}\) is the current position of the ego vehicle, \(x\) is the tensor connected by the output of the feature fusion network and the target point of the ego vehicle.

\begin{equation}
    h_{t}=tanh(W_{hh}h_{t-1}+W_{hx}x_{h}+B_{h})
\label{eq:1}
\end{equation}

The output of the RNN is a sequence of Way points, denoted as \(Wp_{t}\). They are calculated using Equation 2.

\begin{equation}
   Wp_{t}=W_{hy}h_{t}+B_{y}
\label{eq:2}
\end{equation}

The RNN cell used in the model is the Gated Recurrent Units (GRU). By repeatedly stacking these GRU cells, an RNN network can be obtained. In this work, the output \(h_{t}\) of each cell is dynamic. The order of the generated way points is from the target point to the current position of the ego vehicle, as shown in Figure 5.

\subsection{Control Module}

The aim of the control module is to generate actions from the trajectory. The control module will output the steering wheel angle, as well as the use of the throttle. The process is formalized in Equation 3 to Equation 5.

\begin{equation}
  P=\frac{1}{K}\sum W_{k}
\label{eq:3}
\end{equation}

\begin{equation}
   v=\frac{1}{K}\sum\frac{L_{2}(w_{k}-w_{k-1})}{\Delta t}
\label{eq:4}
\end{equation}

\begin{equation}
   \delta=tan^{-1}(\frac{p_{y}}{p_{x}})
\label{eq:5}
\end{equation}

\(P\) is the midpoint of those path points; \(v\) is the desired speed; \(\Delta t\) is the Time interval ; \(\delta\) is the angle between the head of the vehicle ego and the point \(P\). Then, \(v\) will be mapped to the throttle, and the \(\delta\) will be mapped to the steer. The steering wheel is in [-1, 1], where -1 represents the maximum angle of left turn and 1 represents the maximum angle of right turn. The throttle is also in [-1, 1], where -1 is the maximum value

\begin{figure}[htbp]
    \centering
    \includegraphics[width=0.5\textwidth]{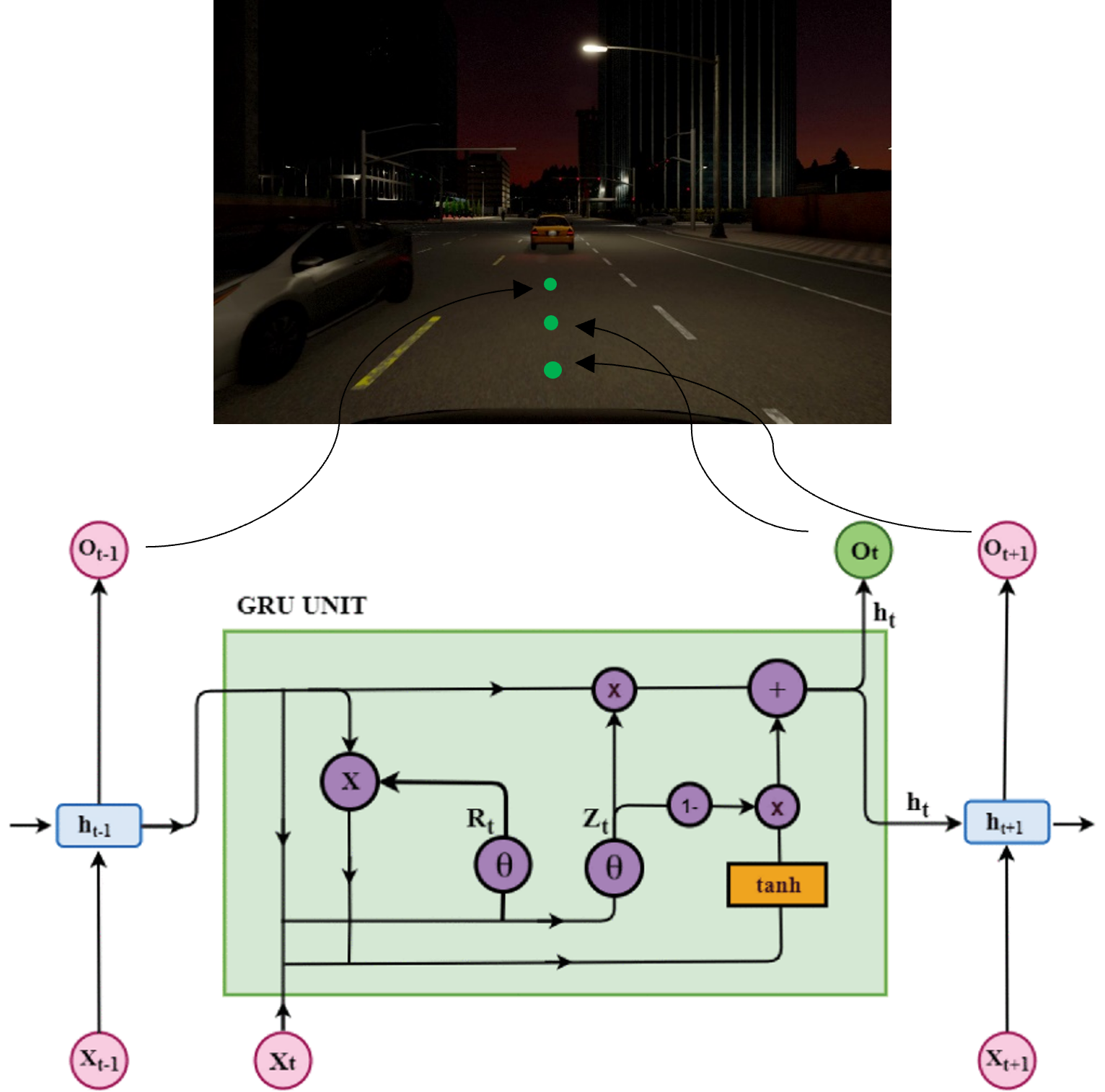}
    \caption{The GRU cells network for planner module}
    \label{fig:5}
\end{figure}

for braking and 1 is the maximum value for acceleration. The process is formalized in Equation 6 and Equation 7:

\begin{equation}
   \mathrm{steer}=\mathrm{Lateral\_PID}(\delta)
\label{eq:6}
\end{equation}

\begin{equation}
  \mathrm{throttle}=\mathrm{Longitudinal\_PID}(v)
\label{eq:7}
\end{equation}

\subsection{Reinforcement Learning Expert}

The training data is collected by an informed reinforcement leaning expert called Roach \cite{zhang2021end}. This paper uses the pretrained Roach agent as the demonstrator for the imitation learning agent. The expert is informed because takes the real-time ground truth BEV map from the CARLA simulator as its input, and generates the control actions through a Proximal Policy Optimization (PPO) \cite{schulman2017proximal} neural network.

The dataset is generated on all routes and scenarios provided by the CARLA leaderboard training suite. The training kit has five maps named, Town01, Town03, Town04, Town06, Town08 and a hundred routes in total within those maps. The data collection process is to let experts go through all routes and scenarios while sampling key information. The frequency of data sampling is 2 frame per second. Considering the generality of the dataset, more information is collected than is required for this model.

\section{Experiments}
\subsection{Implementation}

The model is trained on a GeForce RTX 3090 ti GPU with 24 GB of video memory. The simulator is the released version of CARLA 0.9.10.1. The hyperparameters used in the final model are:

\begin{itemize}
\item The number of objects to detect in an image is 100.
\item The number of GRU cells are 4.
\item The input image resolution is 256 by 256.
\item The learning rate for training the network is \(1e^{-4}\).
\item The number of epochs for training the network is 501.
\item The batch size for traning the network is 128.
\end{itemize}

In addition to the network itself needs to be tuned, the controller parameters of the vehicle are also critical to the final performance of the model. The parameters of the vehicle controller include the following:

\begin{itemize}
\item The lateral PID controller: \(K_{P}=1.25\), \(K_{I}=0.75\), \(K_{D}=0.3\), \(n=30\)
\item The longitudinal pid controller: \(K_{P}=5.0\), \(K_{I}=0.5\), \(K_{D}=1.0\), \(n=40\).
\item The max throttle is 0.75.
\item The maximum value of acceleration is 0.2 \(m/s^{2}\).
\end{itemize}

\subsection{Training Methods}

The data used in the training process of this work are the speed of the vehicle, the image of the front camera, goal point, and way points. The training program uses stochastic gradient descent as the training method and Adam as its optimizer. The loss function is \(L_{1}=\sum_{k}(|\hat{W}_{px}-W_{Px}|+|\hat{W}_{py}-W_{Py}|)\).

During the training process of imitation learning, the imitation learning agent will first use pre-collected data for offline learning. Offline learning does not require the agent to interact with the environment in the simulator. It updates the model parameters through labels of dataset. After 500 episodes of offline training, the model will fit a policy similar to the expert. However, the generality of this policy is not good enough. The ego vehicle will perform badly, if it encounters scenes it has not seen in the training set. Therefore, the data aggregation method is applied to solve this problem. After the offline training of N sets is completed, the expert roach will intervene in the training to provide the imitation learning agent with on policy annotations. During this process, the imitation learning agent will run its policy in the Carla simulator, and at the same time, Roach will get the ground truth information of the current state from the simulator, and make decisions based on it. The information produced by the imitation learning agent and the Roach expert in this process will be recorded as a new dataset. Afterwards, the data of these two datasets will be mixed half and half to form a new dataset. A new round of offline learning will be based on this new dataset to generate a new driving policy. After repeating the above steps 12 times, a robust driving policy is created.

\subsection{Evaluation}

The Detrive model is evaluated on the Carla leaderboard's NoCarsh-dense benchmark. Carla Leaderboard is the most commonly used autonomous driving evaluation system in academia. The competition has been running since 2019 and each year collects entries in cooperation with some international academic conferences. During these years, the benchmark of the competition was updated once in 2020. The benchmark currently used for the competition is the Carla Grand Challenge 2020. The aim of the challenge is to design an autonomous agent to drive the vehicle through a sequence of predefined routes, where each route consists of only a few key waypoints. The agent is required to follow the traffic rules and drive through each key point without collision. The routes in the challenge include cities, villages, and highways in both the United States style and European style. In addition, the challenge also has 14 different weather simulations, such as daylight, sunset, rain, fog, and night. There are 26 routes, and 3 maps in total. The evaluation maps are Town02, Town04, Town06.

The Challenge has designed many common traffic scenarios in its tracks, such as: keep the current lane on congested roads; lane merging on the highway and urban streets; change to the turn lane ahead of the intersection; negotiations at traffic intersections and roundabouts, handling traffic lights and traffic signs, avoid collision with pedestrians and cyclists who suddenly cross the road. It subtracts points for the ego vehicle based on how and how many times it violates traffic rules.

\section{Results}

When evaluated by the CARLA leaderboard, the Detrive model shows higher total score than other imitation learning based models. As shown in Table I, the Detrive model effectively reduced the collisions. However, due to the lack of labeled traffic light, and stop sign data, the model finds it hard to tell if the traffic light is red or green, and if the traffic sign is stop or speed limitation. Therefore, the scores on those metrics are relatively low.

\begin{table}[htbp]
\caption{CARLA LEADERBOARD COMPARISON RESULTS}
\label{tab1}
\begin{tabular}{l|l|l|l|l}
\hline
Model                  & \textbf{OURS}  & WOR \cite{chen2021learning} & LBC \cite{chen2020learning} & CILRS \cite{codevilla2019exploring} \\ \hline
Driving score          & \textbf{34.49} & 31.37        & 8.94         & 5.37           \\
Route completion       & \textbf{67.37} & 57.65        & 17.54        & 14.4           \\
Infraction penalty     & 0.59           & 0.56         & 0.73         & 0.55           \\
Collisions pedestrians & \textbf{0}     & 0.61         & 0            & 2.69           \\
Collisions vehicles    & \textbf{0.32}  & 1.35         & 0.4          & 1.48           \\
Collisions layout      & \textbf{0.49}  & 1.02         & 1.16         & 2.35           \\
Red light infractions  & 0.78           & 0.79         & 0.71         & 1.62           \\
Stop sign infractions  & 0.15           & 0            & 0            & 0              \\
Off-road infractions   & \textbf{0.45}  & 0.96         & 1.52         & 4.55           \\
Route deviations       & \textbf{0}     & 1.69         & 0.03         & 4.14           \\
Route timeouts         & 0.03           & 0            & 0            & 0              \\
Agent blocked          & 1.32           & 0.47         & 4.69         & 4.28           \\ \hline
\end{tabular}
\end{table}

\section{Conclusion}

This paper has proposed a novel end-to-end autonomous driving model that uses a DETR-like structure for perceiving the surrounding driving environment. The overall training method of this work is imitation learning. It contains a informed reinforcement learning expert that can read ground-truth information directly from the simulator, and a student network that can mimic the action of the expert. The focus of this work is to improve the perception module of the imitation learning agent. Instead of the commonly used CNN classification network, a DETR-like detection model is used for perception. Another innovation of this paper is the structure of the feature fusion network. The fusion network, composed of the multi-layer MLP structure and the vector concatenation network can effectively fuse the data of different modalities and reduce the dimensionality. The last part of the Detrive network is a GRU-based RNN, which is used to generate waypoints. Then those waypoints can be mapped into actions by the vehicle control algorithm.

This paper verifies that the use of a transformer-based object detection model end-to-end autonomous driving model can significantly improve the driving performance compared with those classification-based perception models. However, the perception model is only trained with a universal vision dataset, rather than with a dedicated driving scene dataset. This leads to the fact that some traffic objects are not labeled in the dataset, and are therefore hard to be detected by the perception network.

As a future direction to extend our work and to thoroughly evaluate the efficacy of our proposed end-to-end autonomous driving model, we plan to train the proposed model with a driving scene dataset. In addition, we will consider the use of BEV view to help let the vehicle understand the spatial relationship, and therefore resulting in better path planning.

\printbibliography

\end{document}